%% file: main.tex
\documentclass{article} 
\usepackage{colm2024_conference}

\usepackage{microtype}
\usepackage{hyperref}
\usepackage{url}
\usepackage{booktabs}

\usepackage{xcolor}         
\usepackage{xspace}
\usepackage{enumitem}
\usepackage{amsmath}
\DeclareMathOperator*{\argmax}{arg\,max}

\usepackage{amssymb}
\usepackage{mathtools}
\usepackage{amsthm}
\usepackage{multirow}
\usepackage{subcaption}
\usepackage{titlesec}

\newcommand{\model}{JetMoE\xspace}
\newcommand{\lm}{JetMoE-8B\xspace}

\title{\model: Reaching Llama2 Performance with 0.1M Dollars}

\author{Yikang Shen \thanks{Equal contribution.} \\
MIT-IBM Watson AI Lab\\
\texttt{yikang.shn@gmail.com} \\
\And
\hspace{-5cm}Zhen Guo$^*$ \\
\hspace{-5cm}MIT EECS\\
\hspace{-5cm}\texttt{zguo0525@mit.edu} \\
\AND
Tianle Cai \\
Princeton University\\
\texttt{tianle.cai@princeton.edu} \\
\And
\hspace{2.24cm}Zengyi Qin \\
\hspace{2.24cm}MyShell.ai \& MIT\\
\hspace{2.24cm}\texttt{qinzy@mit.edu}
}

\colmfinalcopy 
\begin{document}

\maketitle

\begin{abstract}
Large Language Models (LLMs) have achieved remarkable results, but their increasing resource demand has become a major obstacle to the development of powerful and accessible super-human intelligence. This report introduces \lm, a new LLM trained with less than \$0.1 million, using 1.25T tokens from carefully mixed open-source corpora and 30,000 H100 GPU hours. Despite its low cost, the \lm demonstrates impressive performance, with \lm outperforming the Llama2-7B model and \lm-Chat surpassing the Llama2-13B-Chat model. These results suggest that LLM training can be much more cost-effective than generally thought. \lm is based on an efficient Sparsely-gated Mixture-of-Experts (SMoE) architecture, composed of attention and feedforward experts. Both layers are sparsely activated, allowing \lm to have 8B parameters while only activating 2B for each input token, reducing inference computation by about 70\% compared to Llama2-7B. Moreover, \lm is highly open and academia-friendly, using only public datasets and training code. All training parameters and data mixtures have been detailed in this report to facilitate future efforts in the development of open foundation models. This transparency aims to encourage collaboration and further advancements in the field of accessible and efficient LLMs. The models are publicly available at \url{https://github.com/myshell-ai/JetMoE}.
\end{abstract}

\let\clearpage\relax
\include{intro}
\include{architecture}

\include{data}

\include{training}

\include{alignment}
\include{eval}
\include{limitation}

\section{Conclusion}
We introduce \lm, an open-source MoE model that achieves state-of-the-art performance among open-source models while maintaining high efficiency. By leveraging sparse activation in both the attention and feed-forward layers, \lm reduces computational costs while maintaining strong performance across a wide range of tasks.

Trained using a two-phase approach and a carefully curated mixture of open-source datasets, \lm outperforms larger and more resource-intensive models on the OpenLLM Leaderboard. In addition, \lm-Chat demonstrates competitive performance compared to other open-source chatbots.

We provide detailed training parameters and data mixture information to encourage reproducibility and enable researchers to build upon our work. \lm represents a significant step forward in the development of open-source, efficient, and high-performing language models, contributing to the democratization of advanced language technologies.


\section*{Acknowledgments}
We express our gratitude to Shengding Hu for his valuable advice on the Phase 2 data mixture. We also express our gratitude to Exabits for their assistance in setting up the GPU clusters, and to Lepton AI for their support in setting up the chat demo.

\bibliography{colm2024_conference}
\bibliographystyle{colm2024_conference}

\appendix
\section{Appendix}
You may include other additional sections here.

\end{document}

%% file: intro.tex
\section{Introduction}

Large Language Models (LLMs) have achieved remarkable results, but their increasing resource demand has become a major obstacle to developing powerful and accessible AI. Although modern LLMs have surpassed human performance on some tasks, they remain inefficient and inflexible. Most LLMs (e.g., Llama, \citealt{touvron2023llama}; Pythia, \citealt{biderman2023pythia}; GPT-3, \citealt{brown2020language}; Mistral, \citealt{jiang2023mistral}) use all of their parameters during inference and training, which are referred to as dense models. Considering the substantial costs, the Mixture-of-Experts (MoE) architecture \citep{yuksel2012twenty, shazeer2017outrageously, du2022glam, pan2024dense} has emerged as a popular solution, enabling parameter scaling while keeping computational costs modest. Recent applications of MoE architectures in Transformers \citep{vaswani2017attention} have yielded successful attempts at scaling language models to a substantial size, accompanied by remarkable performance, such as Deepseek MoE~\citep{dai2024deepseekmoe}, Mixtral 8x7B~\citep{jiang2024mixtral}, Grok-1~\citep{grok1_xaiorg}, and DBRX~\citep{dbrx_databricks}. However, even though these models achieve excellent performance, they are not truly open-sourced as the training recipes are not published and may contain proprietary datasets inaccessible outside of large corporations. The open-source community has also attempted to train MoE models, such as OpenMoE~\citep{xue2024openmoe}, but its performance is only on par with weak dense models with similar activation parameters, such as OpenLLaMA~\citep{openlm2023openllama} and TinyLLaMA~\citep{zhang2024tinyllama}.

\begin{figure}[ht!]
    \centering
    \includegraphics[width=0.38\textwidth]{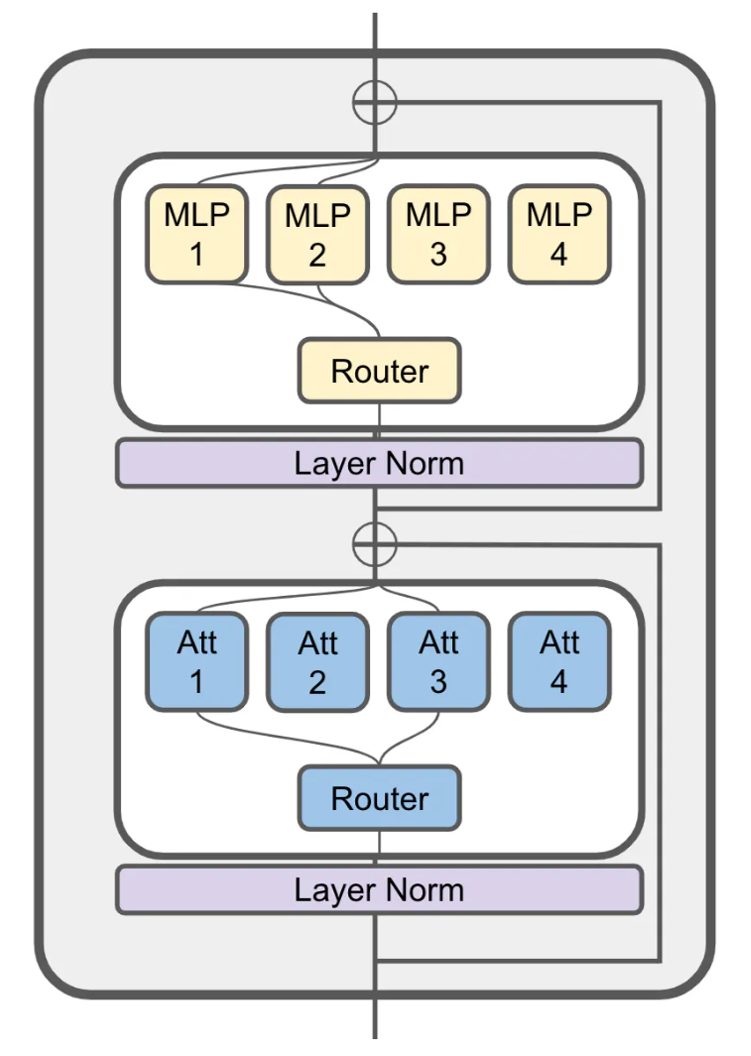}
    \caption{JetMoE architecture}
\end{figure}

To facilitate future efforts on open foundation models, particularly MoE models, we introduce JetMoE-8B, an innovative MoE architecture inspired by ModuleFormer~\citep{shen2023moduleformer} that extends the concept of sparse activation to both the attention and feed-forward layers. Unlike prior works that only apply sparse activation to the feed-forward layer, JetMoE-8B leverages sparse activation in both components to further reduce computational costs while maintaining performance.

Impressively, JetMoE-8B is trained with a limited \$100k budget, using 1.25T tokens from mixed open-source datasets and 30,000 H100 GPU hours. Despite its low cost, JetMoE-8B outperforms the Llama2-7B model, and JetMoE-8B-Chat outperforms the Llama2-13B-Chat model, demonstrating that LLM training can be much more cost-effective than generally thought. In addition, JetMoE-8B has 8B parameters while only activating 2B for each input token, reducing inference computation by about 70\% compared to Llama2-7B.

The key advantages of JetMoE-8B include:

\begin{itemize}[leftmargin=0.85cm]
\item \textbf{Openness and academia-friendly}: JetMoE-8B is trained using only public datasets and open-source training code, making it accessible to many academia research settings. The model can also be finetuned with limited compute budgets (e.g., consumer-grade GPUs).
\item \textbf{Sparse activation on both attention and feed-forward layers}, which significantly reduces training and inference costs. We also propose to share the kv projection in attention experts to improve training stability.
\item \textbf{Comprehensive open-source data mixture}, which ensures high-quality training using only open-source datasets.

\end{itemize}

These innovations in JetMoE-8B pave the way for more accessible and efficient LLMs, benefiting the broader AI research community. To foster collaboration and further advancements, we have detailed all the training parameters and data mixture in this report.

%% file: architecture.tex
\section{Model Architecture}
\subsection{Mixture of Experts}
\newcommand{\x}{\mathbf{x}}
\newcommand{\s}{\mathbf{s}}
\newcommand{\key}{\mathbf{k}}
\newcommand{\val}{\mathbf{v}}
\newcommand{\query}{\mathbf{q}}
\newcommand{\out}{\mathbf{o}}
\newcommand{\W}{\mathbf{W}}
\newcommand{\y}{\mathbf{y}}
\newcommand{\A}{\mathbf{A}}
\newcommand{\B}{\mathbf{B}}

A Mixture of Experts (MoE) layer comprises $N$ modules $f_1,\hdots,f_N$ and a router $g(e \mid \x)$.
Given an input $\x$ to the MoE layer, the router predicts a probability distribution over the $N$ modules.
Of these, we select the top $k$ experts.
When $k < N$ 
, we are using a Sparse Mixture of Experts (SMoE, \citealt{shazeer2017outrageously}). 
In this \model, we use a linear layer to model the router
\begin{align}
    \s &= \W_{rtr} \x \label{eq:gating}, \\
    g(e \mid \x) &= \left\{\begin{matrix}
     \mathrm{softmax} \left( \text{Top}k \left( \s \right) \right)_i, & \s_i \in \text{Top}k \left( \s \right) \\ 
     0, & \s_i \notin \text{Top}k \left( \s \right)
    \end{matrix}\right.
\end{align}
where $\W_{rtr}$ is the expert embedding matrix of shape $(N,D_{\text{emb}})$, $\text{Top}k$ is the operator that select the top $k$ logits from $\s$.
The final output of the SMoE is then given by 
\begin{equation}
    y = \sum_{e=1}^N g(e \mid \x) \cdot f_e(\x)
\end{equation} 
When $g(e \mid \x) = 0$,  $f_e(\x)$ will not need to be evaluated, thus reducing computation cost during training and inference.

Following the design in ModuleFormer~\citep{shen2023moduleformer}, \model replaces both self-attention and Feed-forward layers (FFD) with SMoE layer.
This is different from most opensource MoE models~\citep{dai2024deepseekmoe,xue2024openmoe}, that only replace FFD layers.

\subsection{FeedFoward Expert}
Each FFD expert is a standard 2-layer MLP with hidden state size $D_\mathrm{ffd}$:
\begin{equation}
    f_{mlp}(\x) = \W_{out} \sigma \left( \W_{in} \x \right)
\end{equation}
Where $\W_{out}$ is the output projection matrix of shape $(D_{emb}, D_{ffd})$, $\W_{in}$ in the input projection matrix of shape $(2D_{ffd}, D_{emb})$, $\sigma$ is the SwiGLU activation function.

\subsection{Attention Expert}\label{sec:MoA}

\citet{zhang2022mixture} propose the Mixture of Attention heads (MoA), which extends SMOEs to attention mechanisms.
We adapt MoA for our purposes, generalizing it to allow for multiple heads per expert and introducing RoPE relative positioning into the attention computation.

In \model, each attention expert $e$ is composed of four $\mathbb{R}^{D_{emb} \times D_{att}}$ matrix: $\W_q^e, \W_k, \W_v, \W_o^e$, where $D_{att}=H \times D_{head}$, $H$ is the number of attention head inside each attention experts, $D_{head}$ is the dimension of each attention head. 
Among these matrices, $\W_q^e$ and $\W_o^e$ are owned by each expert, but $\W_k$ and $\W_v$ are shared across experts to improve the training and inference efficiency.

Given an input vector sequence $\x$, we first projected it to key vectors $\key$ and value vectors $\val$ using the shared key and value projection matrices:
\begin{align}
    \key &= \W_{k} \x \\
    \val &= \W_{v} \x 
\end{align}
Inside expert $e$, we project $\x$ into the query vectors $\query_{e}$, apply standard multi-head attention with RoPE~\citep{su2024roformer}, and project the attention output back to the input space:
\begin{align}
    \query_e &= \W_q^e \x \\
    \mathbf{a}_e &= \mathrm{MHA} \left( \query_e, \key, \val \right) \\
    \out_e &= \W_o^e \mathbf{a}
\end{align}
By introducing the MoA, we can scale up the attention layer with more attention experts while maintaining the same amount of computation. 
Such that the attention layer will not become a performance bottleneck, while we scale up the MLP layers.

\subsection{Load Balancing during Pretraining}\label{sec:load_balancing}
To avoid the SMoE repeatedly using the same module and wasting the extra capacity in the other modules, it requires various load balancing losses to regulate the training of the router \citep{shazeer2017outrageously,fedus2021switch}. 
In the training of \model, we use the frequency-based auxiliary loss introduced in \citet{fedus2021switch} 
\begin{equation}
    loss_b = N \sum_{i=1}^N f_i P_i
\end{equation}
where $N$ is the number of experts, $f_i$ is the fraction of tokens dispatched to expert i, and $P_i$ is the fraction of the router probability allocated for expert $i$.
To improve the training stability, we also use the router z-loss introduced in \citet{zoph2022st}:
\begin{equation}
    loss_z = \frac{1}{B} \sum_{i=1}^B \left( \log \sum_{j=1}^N \exp ( x_j^i ) \right)^2
\end{equation}
where $B$ is the number of tokens, $x$ is the logits given by router.
The final training loss will be the weighted sum of three losses:
\begin{equation}
    loss = loss_{lm} + \alpha loss_b + \beta loss_z
\end{equation}
where $\alpha$ is the weight for load balancing loss and $\beta$ is the weight for z-loss.

%% file: data.tex
\section{Pretraining Datasets}
\subsection{Real-world Datasets}

\paragraph{RefinedWeb} is a high-quality web dataset, which contains 5 trillion tokens extracted from CommonCrawl~\footnote{http://commoncrawl.org/} using the MacroData Refinement (MDR) pipeline to improve data quality~\citep{penedo2023refinedweb}. We use the 600 billion token extract of RefinedWeb publicly available.

\paragraph{StarCoder} training data is sourced from The Stack v1.2 with code from GitHub spanning 86 programming languages~\citep{li2023starcoder}. The data is preprocessed through visual inspection, filtering, deduplication, and reweighting low-data languages. A new version of the dataset has been recently released~\citep{lozhkov2024starcoder}.

\paragraph{Dolma} is a large, open, diverse English text corpus contains 3 trillion tokens sampled from 7 sources, including web pages from Common Crawl, code from The Stack, curated web data from C4~\citep{2020t5}, social media conversations from Reddit, academic papers from PeS2o, public domain books from Project Gutenberg, and encyclopedic content from Wikipedia and Wikibooks~\citep{soldaini2024dolma}.

\paragraph{The Pile} is an 825 GB open-source English text corpus for training large language models~\citep{gao2020pile}. It includes 22 diverse, publicly available datasets such as Wikipedia, NIH exPORTER, ArXiv, Books3, BookCorpus2, OpenSubtitles, YTSubtitles, and Enron Emails.

\subsubsection{Miscellaneous}
\begin{list}{$\circ$}{\leftmargin=2.5em \itemindent=0em}
    \item \textbf{Proof-Pile-2} is a 55 billion token dataset of mathematical and scientific documents~\citep{azerbayev2023llemma}. We use the algebraic-stack (11B tokens) subset including numerical computing, computer algebra, and formal mathematics.
    \item \textbf{OpenWebMath} is a large, high-quality, open dataset containing 14.7 billion tokens of English mathematical web text~\citep{paster2023openwebmath}.
    \item \textbf{StackMathQA} is a meticulously curated collection of 2 million mathematical questions and answers, sourced from various Stack Exchange sites~\citep{stackmathqa2024}.
    \item \textbf{OpenAssistant} is a human-generated, human-annotated assistant-style conversation corpus in 35 different languages. The corpus is a product of a worldwide crowd-sourcing effort involving over 13,500 volunteers~\citep{laion_ai_open_assistant_2023}.
    \item \textbf{xP3x} (Crosslingual Public Pool of Prompts eXtended) is a collection of prompts and datasets spanning 277 languages and 16 NLP tasks~\citep{muennighoff2023crosslingual}.
    \item \textbf{CommitPackFT} is a 2GB filtered version of CommitPack to contain only high-quality commit messages on public Github repos that resemble natural language instructions~\citep{muennighoff2023octopack}.
\end{list}

\subsection{Synthetic Datasets}
\paragraph{OpenHermes 2.5} is a large-scale, diverse, high-quality compilation of open-source and custom synthetic datasets~\citep{OpenHermes2.5}. It contains 1 million primarily synthetically generated instruction and chat samples, following a ShareGPT structure. The dataset is compiled from sources including Airoboros 2.2~\citep{durbin_airoboros_2023}, CamelAI domain expert datasets~\citep{li2023camel}, ChatBot Arena (GPT-4 Only)~\citep{zheng2024lmsyschat1m}, Collective Cognition (09-11-2023)~\citep{CollectiveCognition}, CoT Alpaca GPT4~\citep{si2023empirical}, Evol Instruct 70K and 140K~\citep{xu2023wizardlm}, Glaive Code Assistant~\citep{Glaive-code-assistant}, GPT4-LLM~\citep{peng2023instruction}, GPTeacher~\citep{teknium1_gpteacher_2023}, Medical Tasks~\citep{cogstack_opengpt_2023}, MetaMath 40k~\citep{yu2023metamath}, SlimOrca 550K~\citep{longpre2023flan, mukherjee2023orca, SlimOrca}, Platypus~\citep{platypus2023, lightman2023lets, wang2023scibench}, ShareGPT (GPT4-Only)~\citep{lm_sys_fastchat_2023}, and Unnatural Instructions GPT4~\citep{peng2023instruction}.

\paragraph{UltraTextbooks} is a comprehensive collection of high-quality synthetic and human-written textbooks~\citep{UltraTextbooks}. The composition of the dataset incorporating multiple sources such as \texttt{nampdn-ai/mini-peS2o}, \texttt{open-phi/programming\_books\_llama}, \texttt{open-phi/textbooks}, \texttt{nampdn-ai/tiny-strange-textbooks}, and a select high-quality web collection from \texttt{math-ai/AutoMathText}.

\paragraph{UltraChat 200k} is a filtered subset of the UltraChat dataset, which consists of 1.4M dialogues generated by ChatGPT~\citep{ding2023enhancing, tunstall2023zephyr}. The subset was created by selecting a smaller portion of the data, truecasing the text to fix grammatical errors, and removing dialogues where the assistant inappropriately claims to lack emotions or opinions.

\subsubsection{Miscellaneous}
\begin{list}{$\circ$}{\leftmargin=2.5em \itemindent=0em}
    \item \textbf{TemplateGSM} dataset is a novel and extensive collection containing over 7 million grade school math problems with code solutions and natural language solutions~\citep{zhang2024training}.
    \item \textbf{Magicoder-Evol-110K} and \textbf{Magicoder-OSS-75K} datasets are generated using the OSS-INSTRUCT approach, which leverages a LLM to automatically create new coding problems by drawing inspiration from random code snippets collected from open source projects~\citep{wei2023magicoder}.
    \item \textbf{Evol-Code Alpaca} is an open-sourced implementation of Evol-Instruct adapted for code instructions by streamlining, simplifying, and adding code-specific evolutionary instructions~\citep{luo2023wizardcoder}.
    \item \textbf{Code-290k-ShareGPT} is a dataset in the ShareGPT format, consisting of approximately 290,000 sets of conversations~\citep{Code-290k-ShareGPT}. Code-290k-ShareGPT is built upon the existing datasets \textbf{Python-Code-23k-ShareGPT} and \textbf{Code-74k-ShareGPT}.
\end{list}

%% file: training.tex
\section{Model Pretraining}

\subsection{Infrastructures}
We use Megatron~\citep{shoeybi2019megatron} as the training framework and integrate Megablock~\citep{gale2023megablocks} for MoE support.
We further modified the training framework to support MoA (Section~\ref{sec:MoA}) and z-loss (Section~\ref{sec:load_balancing}).
Against the common practice, we choose the Pipeline parallelism introduced in~\citep{narayanan2021efficient} instead of the expert parallelism for model parallel during training.
This is mainly due to two reasons. 
First, Sparse MoE models usually have a narrower hidden state compared to standard transformer models. Thus, the communication cost for pipeline parallelism is smaller. 
Second, we use the dropless MoE schema introduced in \citet{gale2023megablocks,shen2023moduleformer}, which could cause load unbalance across experts. 
Thus, using expert parallel will cause an unbalanced load across devices and result in inefficient training.
Pipeline parallelism could avoid this slowdown because it computes all the experts inside a layer on the same device.
We conduct training on a cluster containing 12 nodes and 96 H100s. 
Inside each node, gpus are connected via NVLinks. 
Infiniband is used for fast communication between nodes.

\subsection{Hyper-parameters}
\begin{table}[ht]
    \centering
    \begin{tabular}{ccccccccc}
        \toprule
        $P_{total}$ & $P_{active}$ & $n_{layers}$ & $D_{model}$ & $N_{experts}$ & Top-$k$ & $n_{kv\_heads}$ & $D_{head}$ & $D_{mlp}$ \\
        \midrule
        8B & 2B & 24 & 2048 & 8 & 2 & 16 & 128 & 5632 \\
        \bottomrule
    \end{tabular}
    \caption{\lm hyperparameters.}
    \label{tab:hyperparameters}
\end{table}

The hyperparameters of \lm are selected based on the common practice for the 1B transformer language model.
We replace all self-attention and MLP layers in the transformer with MoA and MoE. 
Then, we set the same number of experts to 8 and top-$k$ to 2 for every layer.
Such that the model has approximately two times the computation compared to a 1B model.
Following ST-MoE~\citep{zoph2022st}, the weight for load balancing loss and z-loss is set to 0.01 and 0.001, respectively.
Table~\ref{tab:hyperparameters} shows the key hyperparameters in \lm.

\lm is trained with the AdamW optimizer \citep{loshchilov2017decoupled} with a maximum learning rate of 5e-4 and a batch size of 4M tokens with sequence length of 4096.
We employ the Warmup-Stable-Decay (WSD) learning rate schedule introduced in \citet{minicpm2024}. This learning rate scheduler is divided into three stages: the warmup stage (denoted by W, representing the number of steps at the end of the warmup stage), the stable training stage (denoted by S), and the annealing stage (denoted by D):

\begin{equation}
lr(s)=\begin{cases}\frac{s}{W}*\eta, & s< W\\\eta, & W<s<S\\f(s-S)*\eta, & S<s<S+D\end{cases}
\end{equation}

where $0< f(s-S)\leq 1$ is a decreasing function of $s$, and $\eta$ is the maximum learning rate. In our settings, the warmup stage lasts for 10 billion tokens, and the decay stage spans 250 billion tokens. The initial and final learning rates are set to 10\% of the maximum learning rate. A weight decay of 0.1 and gradient clipping of 1.0 are applied during training.

\subsection{Training Data Mixture}
\lm is trained on 1.25T tokens of primarily English data from web documents, mathematics, and code. Similar to the approach advocated in miniCPM~\citep{minicpm2024} and Gemma~\citep{team2024gemma}, we increase the weight of high-quality data during the learning rate decay phase. The training process is divided into two phases:

\begin{itemize}[leftmargin=0.85cm]
\item \textbf{Phase 1} (warmup and stable learning rate): The dataset includes RefinedWeb, Starcoder, The Pile, peS2o from Dolma, and OpenWebMath.
\item \textbf{Phase 2} (decay learning rate): We include additional high-quality data to further improve the model's performance.
\end{itemize}

\begin{figure}[http!]
    \centering
    \includegraphics[width=0.51\textwidth]{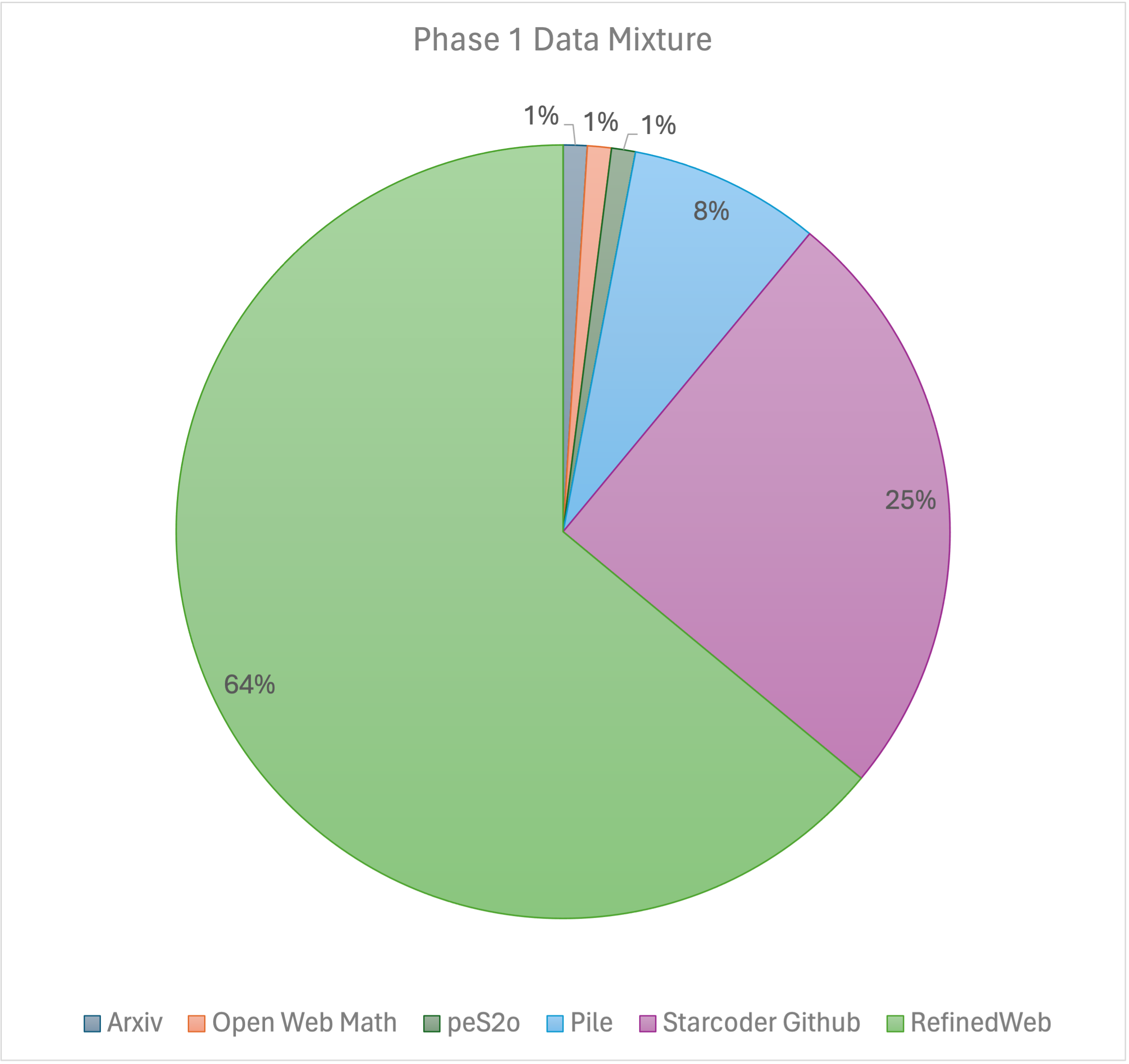}
    \includegraphics[width=0.455\textwidth]{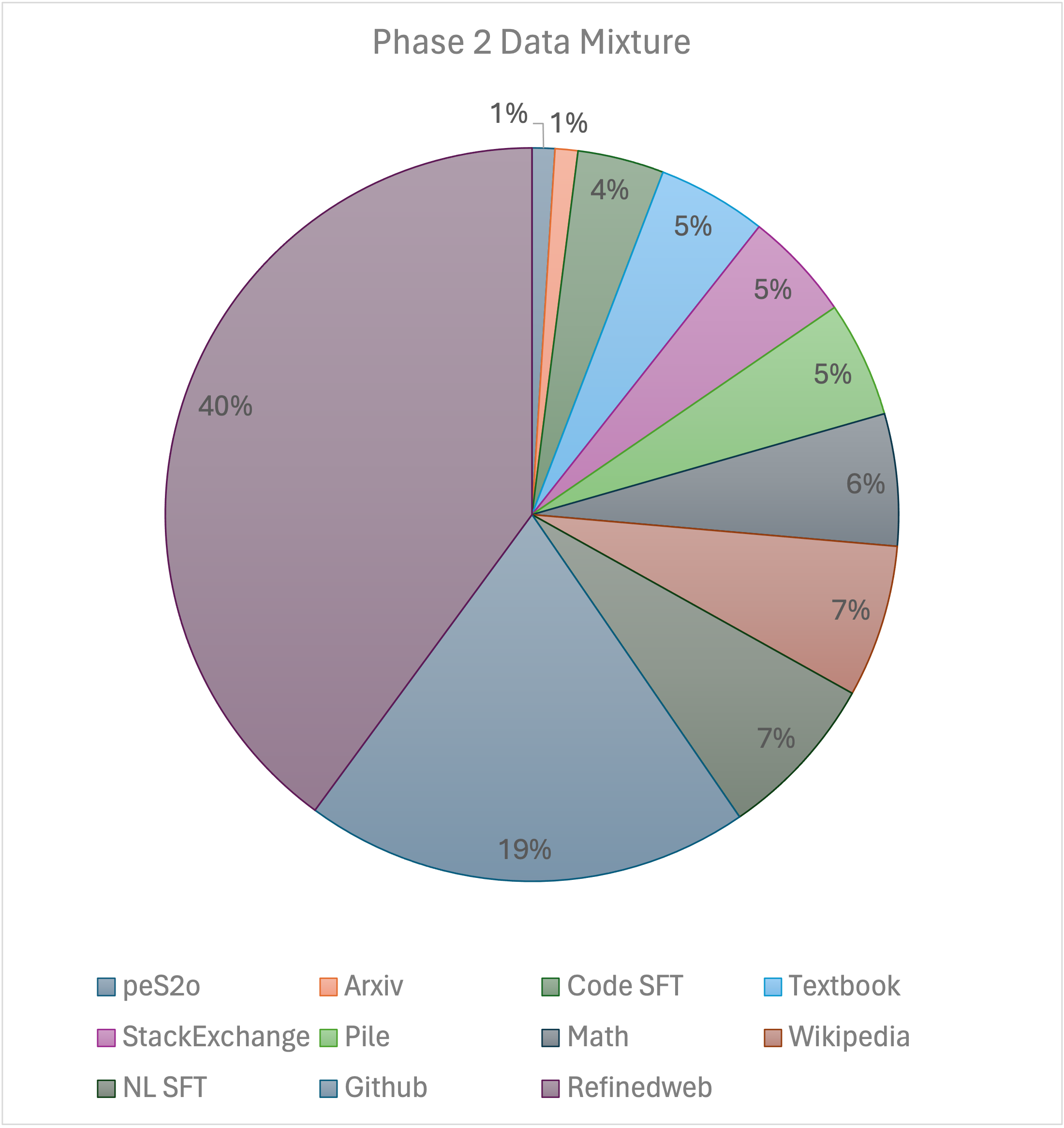}
    \caption{Pretraining data mixture}
    \label{data_mixture}
\end{figure}

The detailed data mixture can be found in Figure~\ref{data_mixture} and Table~\ref{tab:data_mixture}. It is important to note that given the limited computing budget available, our data mixture might not be ideal. However, it serves as a good starting point for training \lm and can be further optimized in future iterations.

\begin{table}[ht]
\centering
\begin{tabular}{llr}
\toprule
\textbf{Category} & \textbf{Dataset} & \textbf{Percentage} \\
\midrule
\multirow{6}{*}{NL pretraining data} & Refinedweb & 39.8\% \\
& Pile\_Wikipedia & 6.7\% \\
& Pile\_StackExchange & 4.8\% \\
& Pile\_arXiv & 1.0\% \\
& Pile\_remaining & 5.1\% \\
& Dolma\_peS2o & 1.0\% \\
\midrule
\multirow{2}{*}{NL SFT data} & xP3x, OpenAssistant, OpenHermes & \multirow{2}{*}{7.3\%} \\
& UltraChat, Oasst-octopack & \\
\midrule
Textbook & UltraTextbooks & 4.8\% \\
\midrule
Code pretraining data & Starcoder Github & 19.6\% \\
\midrule
\multirow{3}{*}{Code SFT data} & Magicoder-OSS, Magicoder-Evol & \multirow{3}{*}{3.8\%} \\
& Code-290k-ShareGPT, CommitPackFT & \\ 
& Evol-Code Alpaca & \\
\midrule
\multirow{2}{*}{Math data} & Open-web-math, algebraic-stack & \multirow{2}{*}{5.8\%} \\
& TemplateGSM, StackMathQA & \\
\bottomrule
\end{tabular}
\caption{Detailed data mixture for Phase 2}
\label{tab:data_mixture}
\end{table}

%% file: alignment.tex
\section{Model Alignment}
\subsection{Distilled Supervised Fine-Tuning (dSFT)}

The dSFT process involves training a student language model for replying to user prompts, with data generated by a teacher model (such as GPT-4 or Claude)~\citep{selfinstruct, alpaca, vicuna2023, tunstall2023zephyr}. The key steps are as follows:

\begin{enumerate}[leftmargin=0.85cm]
    \item \textbf{Data Distillation}: For a set of seed prompts $\{x_j^0\}_{j=1}^J$, generate responses $y_j^0$ using the teacher model $\pi_T$, and refine instructions to obtain $\mathcal{C} = \{(x_j, y_j)\}_{j=1}^J$.
    
    \item \textbf{Instruction Tuning}: The student model $\pi_{\text{dSFT}}$ is trained by maximizing the likelihood of the responses given the instructions:
    
    \begin{equation}
        \pi_{\text{dSFT}} = \argmax_{\pi} \sum_{(x,y) \in \mathcal{C}} \log \pi(y|x).
    \end{equation}
    Note that the expectation for the likelihood function is approximated by using the arithmetic mean over a batch of training samples.
\end{enumerate}

\subsection{Distilled Direct Preference Optimization (dDPO)}

dDPO refines the dSFT model by incorporating preferences from an aligned teacher model into the training process. It optimizes a reward function that reflects these preferences, aiming to align the student model's outputs with the desired outcomes based on the static preference dataset.

\begin{enumerate}[leftmargin=0.85cm]

\item \textbf{KL-Constrained Optimization}: The foundation of dDPO lies in the KL-constrained optimization, which derives the optimal policy $\pi^*_r$ that maximizes expected rewards while minimizing divergence from a baseline policy $\pi_0$~\citep{wang2023reverse}:

\begin{equation}
\pi^*_r(y|x) := \argmax_{\pi} \mathbb{E}_{x \sim d_0} \left[ \mathbb{E}_{y \sim \pi(\cdot|x)}[r(x,y)] - \eta \mathrm{KL}(\pi(\cdot|x)\Vert \pi_0(\cdot|x)) \right]
\end{equation}

where $\eta$ is a regularization parameter that balances maximizing the reward function $r(x,y)$ and adhering to the baseline policy $\pi_0$.

\item \textbf{Preference-Driven Reward Function}: dDPO incorporates a reward function that reflects preferences from an aligned teacher model:

\begin{equation}
r^*(x, y) = \eta \log\left(\frac{\pi^*(y|x)}{\pi_{\text{dSFT}}(y|x)}\right) + \eta \log Z(x),
\end{equation}

quantifying the preference for producing response $y$ given input $x$ relative to the dSFT model's baseline probability. $\eta$ scales the reward's influence, and $Z(x)$ ensures normalization.

\item \textbf{Optimization Objective}: The objective for aligning $\pi_{\theta}$ with the teacher model's preferences is:

\begin{equation}
\pi_{\theta} = \argmax_{\pi} \sum_{(x,y_w,y_l) \in D} \log \sigma \left( \eta \log \frac{\pi(y_w|x)}{\pi_{\text{dSFT}}(y_w|x)} - \eta \log \frac{\pi(y_l|x)}{\pi_{\text{dSFT}}(y_l|x)} \right),
\end{equation}

where $D$ comprises instruction-response pairs, with $y_w$ and $y_l$ indicating preferred and less preferred responses respectively, scored by the teacher model.
\end{enumerate}

Offline DPO~\citep{rafailov2023direct} directly optimizes language model policies using static preference data, providing stable learning and simpler tuning compared to Reinforcement learning from Human Feedback (RLHF)~\citep{ouyang2022training, christiano2023deep}. However, it faces challenges with distribution shifts between the dataset and the evolving policy. Online and iterative DPO variants address this issue at the cost of increased computational complexity~\citep{xu2023things, guo2024direct, xiong2024iterative}.

\subsection{Alignment details}

Our alginment framework is based on Alignment Handbook~\citep{alignment_handbook2023} using Pytorch 2~\citep{he2023transcending, ansel2024pytorch} with DeepSpeed ZeRO-3~\citep{rajbhandari2020zero}. We finetune the JetMoE-8B base model using dSFT on a combination of the following datasets: UltraChat 200k~\citep{ding2023enhancing, tunstall2023zephyr}, Airoboros-3.2~\citep{durbin_airoboros_2023}, Code-Feedback~\citep{zheng2024opencodeinterpreter}, Orca-math-word-problems-200k~\citep{mitra2024orcamath}, SystemChat~\citep{SystemChat}, and Capybara~\citep{daniele2023amplify-instruct}. Chat template is the same as Zephyr-7b-beta. The key hyperparameters for dSFT are a learning rate of 2e-5 with an Adam optimizer, a batch size of 128, and 3 epochs.

We further finetune the JetMoE-8B-SFT model using dDPO on the UltraFeedback dataset~\citep{cui2023ultrafeedback}, which contains binary preference labels indicating the preferred response between two options. The key hyperparameters for dDPO are a learning rate of 5e-7 with AdamW, a batch size of 128, and 1 epoch. This fine-tuning process results in the JetMoE-8B-Chat model. The entire alignment process takes 60 H100 GPU hours.

%% file: eval.tex
\section{Evaluation}
\begin{table}[http!]
    \centering
    \begin{tabular}{ccccccccccc}
        \toprule
        & LLaMA2 & DeepseekMoE & Gemma & \model  \\
        \midrule
        \# Total Params & 7B & 16B & 2B & 8B \\
        \# Activate Params & 7B & 2.8B & 2B & 2.2B \\
        \# Training tokens & 2T & 2T & 2T & 1.25T \\
        \midrule
        ARC-challenge & 53.1 & \textbf{53.2} & 48.4 & 48.7 \\ 
        Hellaswag & 78.6 & 79.8 & 71.8 & \textbf{80.5} \\
        MMLU & 46.9 & 46.3 & 41.8 & \textbf{49.2} \\
        TruthfulQA & 38.8 & 36.1 & 33.1 & \textbf{41.7} \\
        WinoGrande & \textbf{74.0} & 73.7 & 66.3 & 70.2 \\

        GSM8k & 14.5 & 17.3 & 16.9 & \textbf{27.8} \\
        \midrule
        OpenLLM Leaderboard Avg. & 51.0 & 51.1 & 46.4 & \textbf{53.0}\\
        \midrule 
        MBPP (Pass@1) & 20.8 & 34.0 & 28.0 & \textbf{34.2} \\
        HumanEval (Pass@1) & 12.8 & \textbf{25.0} & 24.4 & 14.6 \\
        \midrule
        All Avg. & 45.5 & 47.3 & 43.2 & \textbf{47.6} \\
        \bottomrule
    \end{tabular}
    \caption{OpenLLM leaderboard and code benchmarks results from four different models.} 
    \label{tab:benchs}
\end{table}

We measure \lm's performance on tasks included in OpenLLM leaderboard\footnote{\url{https://huggingface.co/spaces/HuggingFaceH4/open_llm_leaderboard}} and from other domains, including physical reasoning~\citep{bisk2020piqa}, social reasoning~\citep{sap2019socialiqa}, question answering~\citep{clark2019boolq,kwiatkowski2019natural}, mathematics~\citep{cobbe2021training},
commonsense reasoning~\citep{sakaguchi2021winogrande},
language modeling~\citep{paperno2016lambada}, reading comprehension~\citep{joshi2017triviaqa}, and more.
For most benchmarks, we use the same evaluation methodology as in the OpenLLM leaderboard to be comparable to other models.. We compare \lm models to several external open-source (OSS) LLMs, including Gemma, LLaMA2, DeepseekMoE. 

In addition, we include HumanEval~\citep{chen2021codex} and MBPP~\citep{austin2021program} to evaluate the code generation of the models. Utilizing the BigCode Evaluation Harness~\citep{bigcode-evaluation-harness}, we follow recent work on Code LLMs~\citep{2024code, guo2024deepseekcoder} with greedy decoding, and report the mean pass@1 (mean success rate) for the two benchmarks.

Table~\ref{tab:benchs} shows the OpenLLM leaderboard and code benchmarks results from four different models. \lm outperforms Gemma, LLaMA2, and DeepseekMoE on the OpenLLM leaderboard, achieving the best scores in all tasks except ARC-challenge and WinoGrande. Additionally, \lm obtains the highest MBPP scores in Python programming.

\begin{table}[ht]
\centering
\begin{tabular}{lr}
\toprule
\textbf{Model} & \textbf{MT-Bench Score} \\
 \midrule
GPT-4 & 9.014 \\
GPT-3.5-turbo & 7.995 \\
Claude-v1 & 7.923 \\
\textbf{JetMoE-8B-chat} & \textbf{6.681} \\
Llama-2-13b-chat & 6.650 \\
Vicuna-13b-v1.3 & 6.413 \\
Wizardlm-13b & 6.353 \\
Llama-2-7b-chat & 6.269 \\
\bottomrule
\end{tabular}
\caption{MT-Bench score comparison of various models}
\label{table:mt_bench_scores}
\end{table}

\begin{figure}[http!]
    \centering
    \includegraphics[width=0.9\textwidth]{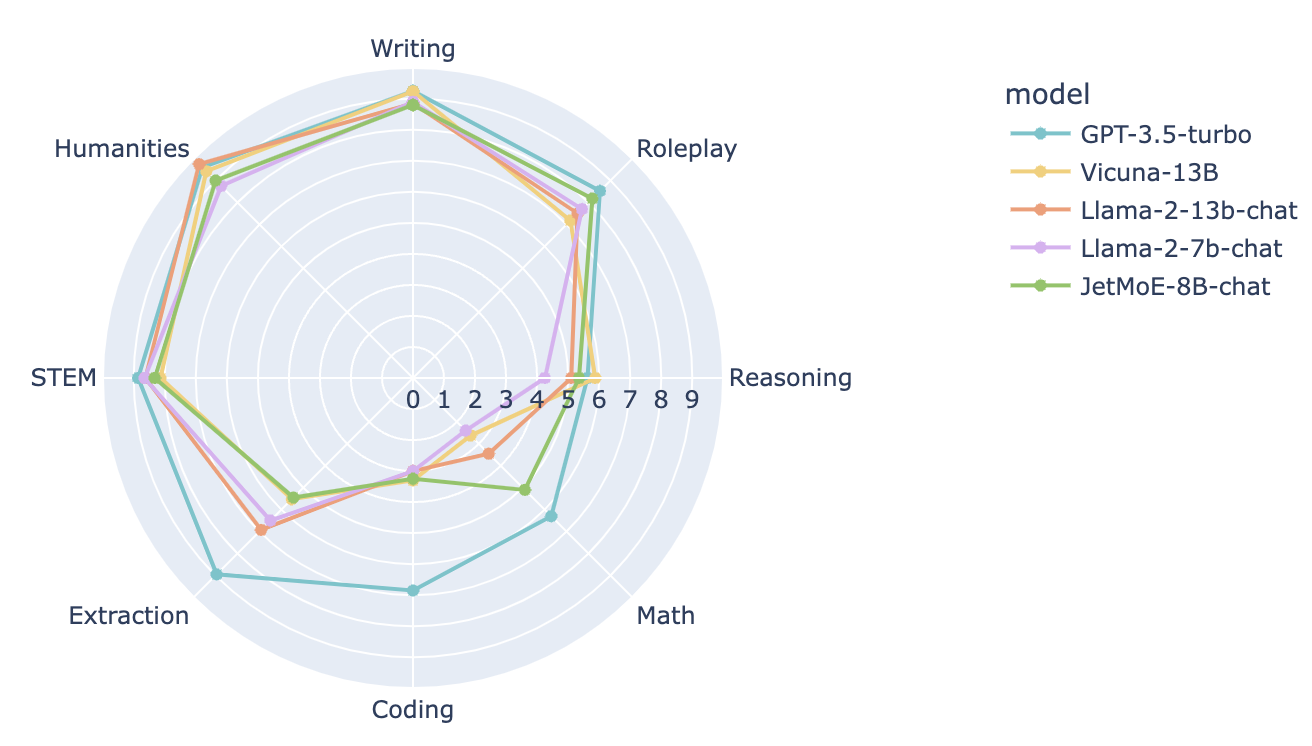}
    \caption{MT-Bench radar figure}
    \label{fig:radr}
\end{figure}

We also evaluated our model on MT-Bench~\citep{zheng2023judging} with a strong LLM judge (gpt-4-0613 checkpoint). The temperature configuration, following the official FastChat implementation, is defined as follows: "Writing" and "Roleplay" tasks have a temperature of 0.7, indicating higher creativity; "Extraction", "Math", "Coding", and "Reasoning" tasks have a temperature of 0.0, suggesting preciseness; and "STEM" and "Humanities" have a temperature of 0.1, implying slightly more variability than 0.0 tasks.

\lm-Chat achieves a higher MT-Bench score than Llama-2-13b-Chat after alignment, demonstrating its superior performance. However, as shown in Figure 3, JetMoE-8B-chat is relatively weak in coding and extraction compared to GPT-3.5-turbo. This might be due to the smaller model size leading to suboptimal reasoning capability in these tasks. Despite this limitation, JetMoE-8B-chat exhibits strong performance across various other dimensions, making it a competitive model in the open-source LLM landscape.

%% file: limitation.tex
\section{Limitation and Future Works}
Due to the limited \$100k budget, we can not afford any ablation study for the model architecture.
The hyperparameters and data mixtures are also handpicked based on the empirical results from previous works~\citep{shen2023moduleformer,zoph2022st,minicpm2024}.
In the future, it would be interesting to further study the actual contribution of different components to the final results.